\documentclass[times,twocolumn,final]{elsarticle}

\usepackage{makecell}
\usepackage[colorlinks=true,linkcolor=black, citecolor=blue, urlcolor=blue]{hyperref}
\usepackage{ycviu}
\usepackage{framed,multirow}
\usepackage{graphicx}
\usepackage{amssymb}
\usepackage{latexsym}
\usepackage{enumitem}
\usepackage{pifont}

\usepackage{tabularx}
\usepackage{url}
\usepackage{xcolor}
\definecolor{newcolor}{rgb}{.8,.349,.1}
\journal{Computer Vision and Image Understanding}

\def\etal{\textit{et al.}~}
\newcommand{\rsec}[1]{Section~\ref{#1}}
\newcommand{\rfig}[1]{Fig.~\ref{#1}}
\newcommand{\rtab}[1]{Table~\ref{#1}}

\newcommand{\new}[1]{#1}

\begin{document}

\ifpreprint
  \setcounter{page}{1}
\else
  \setcounter{page}{1}
\fi

\begin{frontmatter}

\title{Precondition and Effect Reasoning for Action Recognition}

\author[1]{Hongsang \snm{Yoo}} 
\ead{hongsang.yoo@student.unimelb.edu.au}
\author[1]{Haopeng \snm{Li}}
\ead{haopeng.li@student.unimelb.edu.au}
\author[2]{Qiuhong \snm{Ke}\corref{cor1}}
\ead{qiuhong.ke@monash.edu}
\cortext[cor1]{Corresponding author: Qiuhong Ke. Hongsang Yoo and Haopeng Li have equal contributions.}
\author[1]{Liangchen \snm{Liu}}
\ead{liangchen.liu@unimelb.edu.au}
\author[3]{Rui \snm{Zhang}}
\ead{rayteam@yeah.net}

\address[1]{School of Computing and Information Systems, The University of Melbourne, Melbourne, Australia}
\address[2]{Department of Data Science \& AI, Monash University, Melbourne, Australia}
\address[3]{Graduate School at Shenzhen, Tsinghua University, Shenzhen, China}

\received{1 May 2013}
\finalform{10 May 2013}
\accepted{13 May 2013}
\availableonline{15 May 2013}
\communicated{S. Sarkar}

\begin{abstract}
Human action recognition has drawn a lot of attention in the recent years due to the research and application significance. Most existing works on action recognition focus on learning effective spatial-temporal features from videos,  but neglect the strong causal relationship among the precondition, action and effect. Such relationships are also crucial to the accuracy of action recognition. In this paper, we propose to model the causal relationships based on the precondition and effect to improve the performance of action recognition. Specifically, a Cycle-Reasoning model is proposed to capture the causal relationships for action recognition. To this end, we annotate precondition and effect for a  large-scale action dataset. Experimental results show that the proposed Cycle-Reasoning model can effectively reason about the precondition and effect and can enhance action recognition performance.
\end{abstract}


\end{frontmatter}


\section{Introduction}
\label{intro}
Great progress has been made in action recognition due to the success of deep neural networks and multimodal learning in the past few years \citep{action-methods}.
However, little effort has been exerted to focus on the study of interaction between precondition-action and action-effect in this task.
Precondition and effect play an important role in empowering an agent with the ability to reason precondition-action and action-effect relation. 
Despite recent advances in artificial intellegence (AI), artificial agents still lack the ability to understand the relationship between action and effect in the real world \citep{reasoning-nlp}. 
It is critical for artificial agents to be able to understand action and effect relation in order to simulate the action and reason in real world.
For example, it is easy for humans to infer that an action of dropping a glass will lead to the state it is broken or lies on the floor, but for robots, the task still remains challenging.
Furthermore, action recognition could be better generalized in the context of precondition and effect \citep{reasoning-transformation}. 
For instance, in \citep{reasoning-transformation}, it was shown that modelling actions as a transformation from precondition to effect led to stronger generalization in cross-category beyond learned action categories.
Therefore, the importance of roles of precondition and effect in action recognition cannot be emphasized enough.

\begin{figure}[tbp]  
	\centering
	\includegraphics[width=0.48\textwidth]{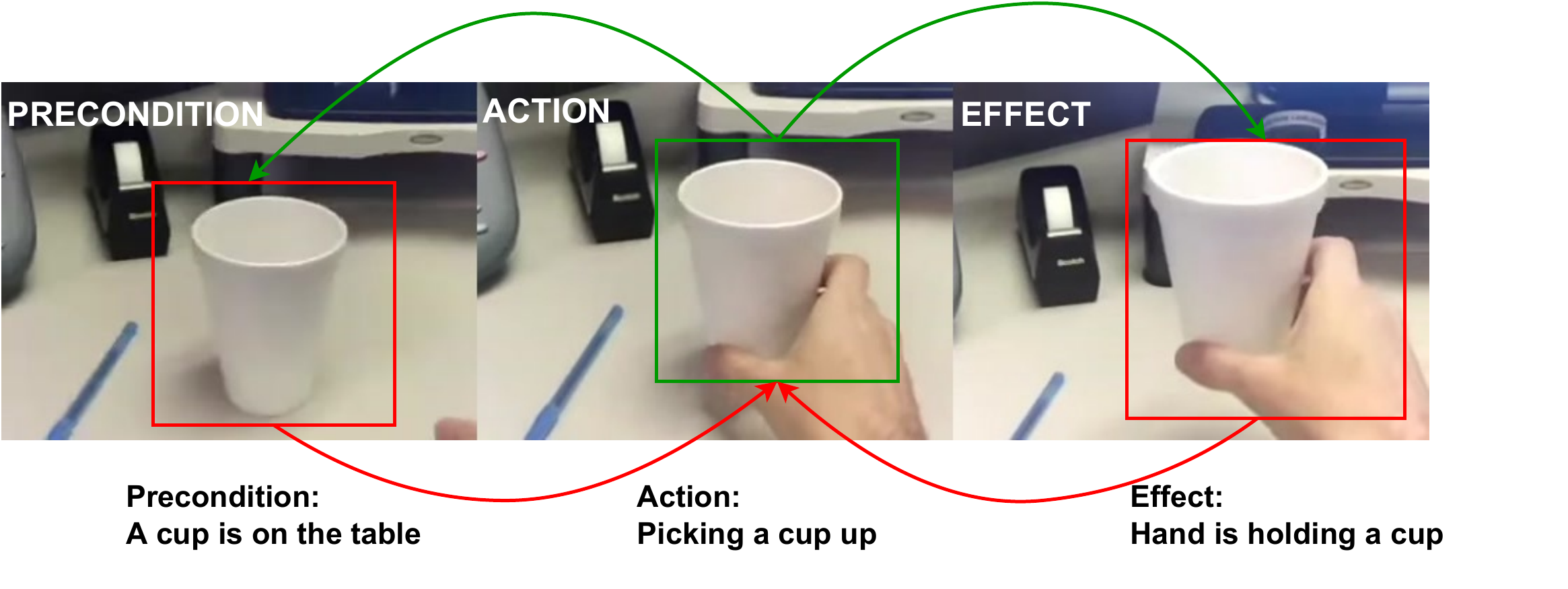}
	\caption{\textbf{Relationship among the precondition, action and effect.} We argue that action can be used to predict precondition and effect. And likewise precondition and effect in turn can be used to improve action recognition}
	\label{figure_main_arrow}
\end{figure}

More specifically, in this context, learning the relationship among precondition, action and effect is meaningful. 
Although the research on precondition, action and effects is prevalent in traditional action planning \citep{reasoning-planning, reasoning-pddl, reasoning-strips, reasoning-context}, modelling their relationships in terms of action recognition is under-explored.
Such study could be beneficial for both precondition and effect task and action recognition task.
For example, on the one hand, side information such as precondition and effect can be an informative indicator to action as it provides extra information which cannot be directly extracted from visual features.
On the other hand, action information can also be beneficial for the task of precondition and effect as it provides counter part for causal relationship features.
This interaction among precondition-action and action-effect is illustrated in~\rfig{figure_main_arrow}.

Although exploring the benefit by learning the relationship among the precondition, action and effect is promising, there exists no dataset containing such information.
To the best of our knowledge, there were attempts \citep{reasoning-Explainable} to study this relationship by re-annotating existing datasets for action recognition. However, these re-annotated datasets only included some features for action recognition task in the context of precondition and effect, but did not contain labels of precondition or effect. 
The lack of precondition and effect labels limits its ability to achieve the goal of precondition and effect reasoning for action recognition. 
Therefore, it is critical to construct a dataset with these dedicated annotations as the test base for this scenario.

In this paper, we target at this, and build a novel annotation set by identifying and performing annotation of meaningful concepts namely precondition, effect and super-class, which are closely related to actions, and develop several novel models to utilize the annotation to predict action and effect.
Our model learns the precondition/effect class based on the probability distribution of action, and learns the action class based on probability distribution of precondition/effect.

The main contributions are summarized as follows:
\begin{enumerate}
    \item Based on the action labels from the dataset, we newly created labels relevant to action (i.e., precondition, effect) to infer their correlations.
    \item We developed several models by leveraging these annotations as a feature and concatenating it with visual features.
    \item We showed that not only these annotations contribute to the significant improvement of action recognition, but also that performance can be continuously improved in cycles where annotation class and action class keep increasing the other's accuracy.
\end{enumerate}

The rest of this paper is organized as follows. We will review related works for this project in \rsec{Related}. Next, in \rsec{Dataset}, we will introduce datasets used in our work and discuss how we created labels of precondition, effect and super-class. Then, in \rsec{Methodology}, we will discuss how we developed the models\footnote{Code and models are available at: \url{https://github.com/kaiyoo/Precondition-and-Effect-Reasoning-for-Action-Recognition}} applying these labels. In \rsec{Experiments}, we will compare the experiment results and conduct analysis on the relationship of action and annotated labels. Finally, we conclude our works and summarize our contribution.

\section{Related work}
\label{Related}
We review the \new{related work from two aspects}: action recognition and relationship reasoning in action. 

\subsection{Video-based Action Recognition}
In the AI planning area, an action can be represented by the semantic level state transition from precondition to effect \citep{reasoning-planning}. 
Therefore, action is defined as a process of transformation that lies between the state of precondition and effect.

There has been a great deal of work \citep{slowfast,nonlocal,spatiotemporal,i3d,c3d,TSM,TPN} in video action recognition and prediction.
In action recognition, the model recognizes a human action from a complete video containing full action execution whereas in action prediction, the model infers a human action from temporally incomplete video data \citep{recognition-prediction}.
Thus, the key difference between the two tasks lies in the time of decision-making, in other words, when to make the inference. 
In this work, we focus on the task of RGB-based action recognition since it is the mainstream methods on the task of video action recognition.
Two popular methods of RGB-based action recognition are Two-Stream 2D CNN-Based Methods and 3D CNN-Based Methods~\citep{action-methods}.

\subsubsection{RGB-based Action Recognition Methods}
Generally, two-stream 2D CNN methods contain two separate CNN branches taking different input features extracted from the RGB videos and use the fusion strategies \citep{action-methods} introduced from popular 2D CNN-based networks in \citep{2D-spatiotemporal-multiplier, 2D-Conv-Fusion, 2D-large-scale-CNN, 2D-2STREAM-CNN, 2D-dynamic}.
Among many works, a classic two-stream CNN model proposed by Simonyan and Zisserman~\citep{classic-2dcnn-method} consists of a spatial network and a temporal network to which RGB frames and optical flows are fed.
These two streams learn appearance and motion features and are fused to produce the final result.
Other variants of two-stream  2D  CNN  methods include TSN (Temporal segment networks)~\citep{TSN}, TSM (Temporal shift module)~\citep{TSM}, TRN (Temporal relational reasoning)~\citep{TRN}. 
Deployment of separate streams in these methods limits interaction between semantics of input frames in the early stage.

On the other hand, 3D CNN-Based Methods \citep{3D-spatiotemporal, 3D-CNN, 3D-fewshot, 3D-directional-tempo} have applied 3D CNNs to simultaneously model the spatial and temporal semantics in videos. Ji et al.~\citep{early-3dcnn-method} proposed one of the earliest 3D CNN-Based methods, which applies 3D convolutions and  produces the final feature representation combining information from multiple channels to extract features from the spatial and temporal dimensions.
Another 3D CNN-Based Method includes C3D network proposed by Tran et al.~\citep{c3d}, which leverages deep 3-dimensional convolutional networks trained on a large-scale supervised video dataset to learn the spatio-temporal features. 
However, the C3D network limits its ability to model spatio-temporal information effectively due to its relatively shallow network.

\subsubsection{Various Visual Tempo Modeling}
Methods discussed above have the limit of having the large variation of visual tempos, which represents how an action goes in the temporal scale.
Visual tempo has shown its benefits in the task of action recognition \citep{slowfast}. 
Due to its nature in the sequence of frames, many previous works such as \citep{slowfast, TPN} suggested network structure modeling such visual tempos of different actions in order to deal with change in information over frames.

Other work targeting on complex visual tempos includes~\citep{slowfast, DTPN},
SlowFast~\citep{slowfast} and DTPN~\citep{DTPN} which construct input-level pyramids using hard-code schemes. However, they are hard to scale up and require multiple frames. 
Similar drawbacks appear on other variants of Two-Stream 2D CNN-Based Methods such as TSN, TSM, TRN~\citep{TSN,TSM,TRN}. 
Furthermore, in these networks, the dynamics of visual tempos are hard to be captured because feature extractions were mainly done in input levels.
This makes the context information of the input frames harder to interact with other features in the early stage and require a multi-branch network to handle.

\subsubsection{Feature-level Visual Tempo Modeling}
To tackle this challenge, Yang et al.~\citep{TPN} proposed a Temporal Pyramid Network (TPN) at the feature-level. 
There were early approaches to construct feature-level pyramid networks \citep{feat-lev-pyr-hypercols,  feat-lev-pyr-interwiner, feat-lev-pyr-network, feat-lev-pyr-path-aggr} to handle the large variance of spatial scales.
Divergent from them, TPN deals with various visual tempos by constructing feature hierarchy. 
TPN has the architecture that can be integrated into 2D or 3D backbone networks using ResNet \citep{resnet-50} or inflated ResNet \citep{slowfast} respectively in a plug-and-play manner.
The variants of TPN construct the hierarchy by utilizing the different source of features and the fusion of features in order to extract features at various tempos. 

\subsection{Relationship Reasoning in Action}
\label{Related.B}
\subsubsection{Precondition, Action and Effect in Planning}
The precondition is the state that must be made true before action execution. On the other hand, the effect is the state achieved after action execution~\citep{reasoning-Explainable}. Therefore, modelling human behaviour in terms of preconditions and effects seems a natural way of capturing dynamics of actions. However, the study on the precondition and effect has not yet been fully investigated in the environment of action recognition task. 

In the AI planning problems domain, an action can be represented by the semantic level state transition from precondition to effect~\citep{reasoning-planning}. In this planning domain, Planning Domain Definition Language (PDDL) is used to describe actions as abstract operators based on preconditions and effects which can be parameterized with domain specific elements~\citep{reasoning-pddl}.

\subsubsection{Action Recognition with Precondition and Effect}
\label{Related.B.2}
As one of earlier works that attempts action recognition in the context of precondition and effect, Yordanova et al.~\citep{reasoning-context} proposed an approach for activity recognition where they modelled human behaviour based on preconditions and effects with PDDL and transformed the causal model into a probabilistic model.
However, it is harder to infer actions with this approach in complex real world scenarios that need models with huge state-space, and most of modern action recognition methods are based on deep learning.

Zhuo~\etal\citep{reasoning-Explainable} proposed action reasoning framework that uses prior knowledge to explain semantic-level observations of state changes by both classical reasoning and deep learning approaches. They construct a scene graph from a video sequence on each frame to represent relevant objects, attributes, and relationships and to track changes of semantic-level states. Their approach of representing objects, attributes, and relationships was partially adopted in the annotation of labels of our work. Wang~\etal\citep{reasoning-transformation} proposed a network by modelling an action as a transformation from precondition to effect on a high-level feature space. The 43 action classes in the dataset used in their experiment were further grouped into 16 \new{super-classes, which mean higher level of action categories. The same action under different subjects, objects and scenes form sub-categories of super-classes.}
Interestingly, their approach of transforming actions from precondition to effect not only showed improvement of action recognition task, but also robust generalization beyond learned action categories and cross-category. 
Their results show that precondition and effect could play an important role in action recognition.
In addition, the fact that their model showed better generalization beyond action category indicates that super-class could act as a useful feature to be incorporated along with precondition and effect. Motivated by this idea, we also included the concept of super-class in our work for the relationship reasoning in action.

\subsubsection{Action and Effect Prediction}
Gao~\etal\citep{reasoning-nlp} addressed the relations between actions and effects on the state of the physical world using images. They used web image data through distant supervision for action-effect prediction.Their work showed that the use of effect descriptions led to better performance for action and effect prediction.
Although this research used static images rather than video sequences and is more NLP related, it directly targeted prediction of effect as well as action, while the works~\citep{reasoning-Explainable, reasoning-transformation} above focused on action recognition in the relationship of precondition and effects. 


\section{Dataset} 
\label{Dataset}
\new{Although some works on modelling precondition and effect in terms of action recognition was introduced in~\rsec{Related}, there is still a lack of annotation of labels for precondition and effect.}
In this section, we build a novel annotation set consisting of precondition, effect and super-class based on the existing action classes to observe their correlations with action.
The new annotation set is built based on something-something v2 dataset \citep{sth-sth} by identifying and performing annotation of meaningful concepts namely precondition, effect and super-class closely related to actions.
We will discuss the profile of the dataset, its action labels, and the process of creating new labels of precondition, effect and super-class based on action labels. 
 
\subsection{Data Source Profile}

Something-something v2 dataset~\citep{sth-sth} is a large collection of video clips where people perform basic actions with basic objects.
It is created by a large number of crowd-sourced workers.
The dataset contains 220k videos, which is the number more
than twice of its version 1, and the resolution of the video is \new{ 240 × 240 pixels}.
The number of instances for training/validation/testing is 168,993/24,557/27,157.

\begin{figure}[tbp]
	\centering
	\includegraphics[width=0.48\textwidth]{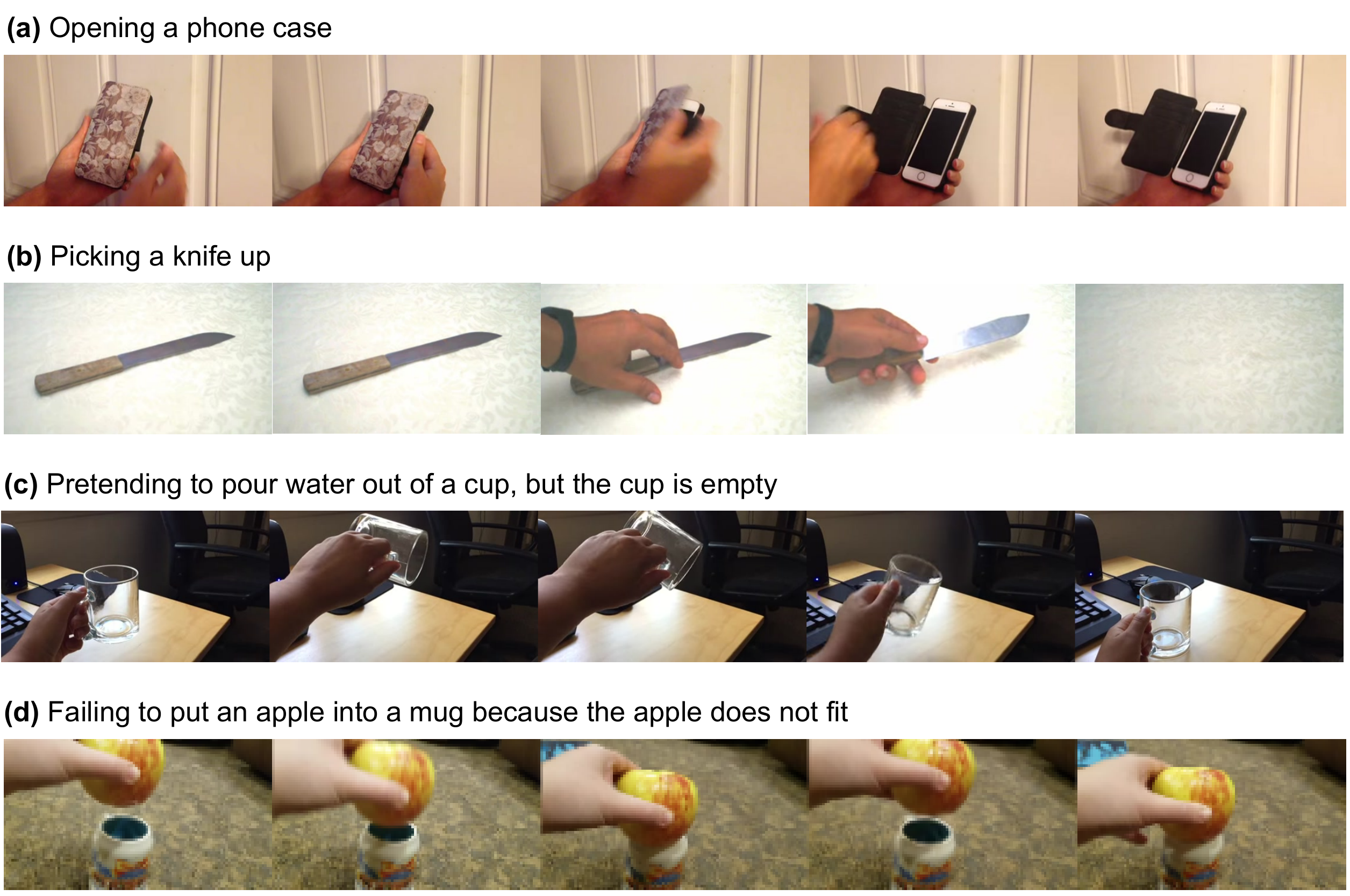}
	\caption{\textbf{Examples of videos and labels. } (a) and (b) are examples of simple actions. (c) and (d) show more complex actions, where some action is performed first and then taken back at a later time.}\label{video_label}
\end{figure}

Among the data above, we used the training and validation data. We did not use the test data since our goal is to show and compare accuracy by different models, for which purpose, labels are required.
Some basic example of these action labels are shown in (a), (b) of~\rfig{video_label}: 

\begin{itemize}
\item \texttt{Opening something}
\item \texttt{Picking something up}
\end{itemize}

These simple actions are easy to recognize. However, there are also some tricky action labels. In~\rfig{video_label}, (c) and (d) are examples of more complex actions where people in the video first perform some actions and then take the action back or fail to complete the intended actions such as:

\begin{itemize}
\item \texttt{Pretending to pour something out of something, but something is empty}
\item \texttt{Failing to put something into something because something does not fit}
\end{itemize}

Another example of action labels will be introduced in the following sections regarding the process of annotations of precondition, effect and super-class.

\subsection{Annotation of Labels Relevant to Action}
\label{Dataset.B}
To extend the original 174 action labels in the original something-something dataset, we identify and manually create labels of precondition, effect and super-class by human insights, which are also closely relevant to the action class.
\new{
Examples of annotations for different action classes are shown in~\rtab{table:labels_table}.} 

  

\begin{table}[tbp]
\new{
\caption{\textbf{Comparison between labels.} Action is the label for existing dataset and the others (precondition, effect and super-class) are newly annotated labels for this work. The number in parenthesis next to each label is its total number of classes.}\label{table:labels_table}
}
  \begin{center}
  \renewcommand{\arraystretch}{1.1}
  
  \renewcommand\theadalign{bc}
  \renewcommand\theadgape{\Gape[4pt]}
  \renewcommand\cellgape{\Gape[4pt]}

  \resizebox{\columnwidth}{!}{%
  
  \begin{tabular}{cccc}

    \hline
    Action (174) & Precondition (60) & Effect (88) & Super-class (23) \\
    \hline
    opening something & something is closed & something is open & appear \\
    closing something & something is open & something is closed & hide \\
    \makecell{Pretending to close something \\ without actually closing it} & something is open & something is open & pretend \\
    Picking something up & something is on the table & hand is holding something & take \\
    Pushing something so it spins & \makecell{Hand is touching something \\ AND something is on the table} & something is spinning  & spin \\
    \makecell{Moving something across \\a surface until it falls down} & something is on the table & something fell off the table & move \\
    Piling something up & something is piled up & something is piled up & put \\

    \hline
  \end{tabular}%
}
  \end{center}
\end{table}

\new{To annotate three new types of labels}, one can follow the annotation method as for the something-something dataset, the labels of which are annotated by a large number of crowd workers for every instance. 
However, this method is not efficient and labour intensive. 
\new{To that end, we conduct the annotation work by human inference from the action labels. In other words, annotated labels were created based on the action class, not on the whole instances. 
This means that an instance with a certain action class will always have the same class for precondition, effect and super-class respectively. 
Therefore, we clarify that our newly created annotation set are not ground-truth labels, but we will see how these pseudo-labels contribute to increasing performance of action recognition in the later section.
Due to the subjective nature of the annotation work, we follow the guides of referring to several videos for all action classes to avoid the subjective bias.
}

In the process of annotations, we represented a label as the combination of smaller propositions (atom), following the annotation methods of Zhuo et al. \citep{reasoning-Explainable}. In their work, they represented action as a set of features in that in the real world, a complex action can be divided into a set of atomic propositions such as attribute and relationship transitions.

Overall, the work of creating labels consists of two stages: 1) creating atoms and 2) creating each label by concatenating one or more atoms. 
First, we divided the descriptive fact about instances to atomic propositions (atom) under different categories.
For example, in the case of effects as shown in~\rtab{table:atom_example_effect}, there are six categories:
hand, camera, location, attribute, relationship between objects and third object.
Then, each label is created by concatenating one or more atoms.
Precondition has different atoms from effect, but the same idea is applied on the process of creation of atoms.

\begin{table}[tbp]
   \caption{\textbf{Examples of atoms and category of effect labels. } A Label consists of one or more of these atoms.}\label{table:atom_example_effect}
 \begin{center}
   \renewcommand{\arraystretch}{1.1}
  \begin{tabular}{ll}
    \hline
    Atom Category & Examples \\
    \hline
    \multirow{3}{*}{hand} & hand is holding X  \\ 
    & hand is not holding X \\ 
    & hand is touching X \\ 
    \hline
    \multirow{2}{*}{camera} & X is in close-up view  \\ 
    & X is at bottom side of camera angle \\ 
    \hline
    \multirow{3}{*}{location} & X is on the surface \\
    & X is on the slanted surface \\ 
    & X is on the table \\ 
    \hline
    \multirow{3}{*}{attribute} & X has a hole \\ 
    & X is broken \\
    & X is collapsed \\
    \hline
    \multirow{3}{*}{relationship} & X and Y are in contact \\ 
    & hand, X, Y are close \\
    & X is in front of Y \\
    \hline
    third object & table is clear  \\ 
    \hline
  \end{tabular}
  \end{center}
  
\end{table}

We present the process of annotation of precondition, effect and super-class. 
The list of annotated labels is available here\footnote{https://github.com/kaiyoo/Precondition-and-Effect-Reasoning-for-Action-Recognition}.

\subsubsection{Annotation of Preconditions}
\label{annotation_precondition}
For creating precondition labels, we defined the criteria as occurring at 30\% of time range of input frames \new{from the hypothesis and observation that most of actions are likely to happen after that point.}
In other words, we sample the event which happens at around 30\% of time range of input frames to represent precondition before the action is performed.
\new{This criteria of 30\% cannot be generalized to all cases due to subjective human inference as previously mentioned, thus we clarify that this is not the ground-truth precondition.}

\begin{figure}[tbp]
	\centering
	\includegraphics[width=0.5\textwidth]{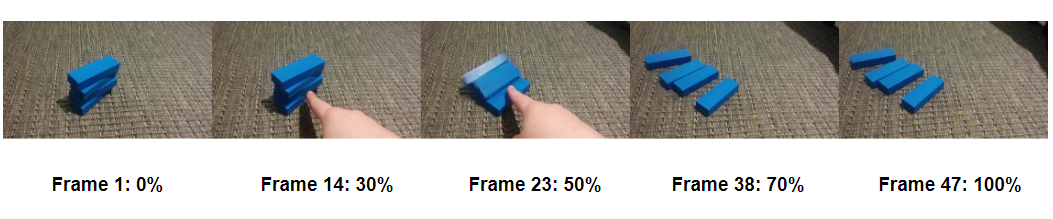}
	\caption{\textbf{Sequence of images at different time-step. } Precondition and effect can be observed at 30\% and 70\% of input frames, respectively}\label{seq_images}
\end{figure}

For example as shown in~\rfig{seq_images}, if the action label is “Poking a stack of something so the stack collapses”, the precondition label would be “finger is seen AND something is piled up” because in order to poke something, in general, a finger can be seen in the frame before the action of poking, which is thought to have started at 30\% of original frames.
\begin{itemize}
\item \textsl{Action}: \texttt{Poking a stack of something so the stack collapses}
\item \textsl{Precondition}: \texttt{finger is seen AND something is piled up}
\end{itemize}
Then, for the clause ``so the stack collapses" to happen, something must have been piled up in the first place. 

In~\rfig{seq_images}, the second image is the 14th frame out of the total 47 input frames, which is about 30\% of time range of the input frames.
Before action of poking the stack of blocks is performed, as shown in the 14th frame in~\rfig{seq_images}, the precondition shows that {\itshape a finger is seen} AND {\itshape something is piled up}.

Precondition label was created by concatenating one or more atoms discussed above by using AND/OR. 
In the example above, “finger is seen” is one atom and “something is piled up” is another atom. 
In this case, the precondition label was created by concatenating these two atoms.

While permutation of the atoms could result in a large amount of labels, most of them are trivial or incompatible (i.e., something is on the table AND the table is clear).
Therefore, we only choose the \new{atoms that can be generalized in most cases intuitively} for the precondition label construction. 
The total number of precondition labels created as such is 60.

\subsubsection{Annotation of Effects}
\label{annotation_effect}
For creating effect labels, we defined the criteria as generalizing what would happen at around 70\% of time range of input frames \new{based on our hypothesis and observation that the effect of performing action is likely to be able to be observed from that point.}
\new{As discussed in~\rsec{annotation_precondition}, we clarify that this criteria of 70\% is not the ground-truth effect due to the subjective human inference.}
For the same example we saw in~\rsec{annotation_precondition}, if the action label is “Poking a stack of something so the stack collapses”, the effect label would be “something is collapsed” because the object is probably collapsed after the action is done and the action is thought to have finished at 70\% of time range of input frames. 
\begin{itemize}
\item \textsl{Action}: \texttt{Poking a stack of something so the stack collapses}
\item \textsl{Effect}: \texttt{something is collapsed}
\end{itemize}
More specifically, as shown in~\rfig{seq_images}, this instance \new{consists of 47 frames.}
Here, the 4th sub-figure shows the 38th frame out of the total 47 frames, which is about 70\% of the input frames. 
After action of doing something is performed, the effect shows that {\itshape the stack is collapsed}.
For those samples whose actions are not clear at 70\% of time range of input frames, we refer to original videos to have better ideas.
In some cases, an action label itself indicates the meaning of effect:
\begin{itemize}
\item \texttt{Pouring something into something until it overflows}
\item \texttt{Moving something and something so they collide with each other}
\item \texttt{Pulling two ends of something so that it separates into two pieces}
\end{itemize}
In this example, the conjunctive adverb such as “until” or “so” and subordinating conjunction such as “so that” reflect what would happen after the action is performed. In this case, we can easily infer effects from action labels by itself, so the effect label for above cases respectively would be:
\begin{itemize}
\item \texttt{something is spilled on the surface}
\item \texttt{something and something2 are in contact}
\item \texttt{something is separated into two pieces}
\end{itemize}

Effect label was created by concatenating one or more atoms by using AND/OR. 
In the above example of~\rfig{seq_images}, the effect is “something is collapsed”, and this effect label consists of one atom.
Similar to the precondition labels, we only choose the \new{atoms that can be generalized in most cases intuitively} for the effect label.
The total number of created effect labels is 88.

\subsubsection{Annotation of Super-class}
Following the super-class concept from~\citep{reasoning-transformation} as discussed in~\rsec{Related.B.2}, we also identify and annotate the super-class of the existing action labels in something-something dataset.
Super-class is a concept that groups action level labels, by their similar characteristics into the same category of higher hierarchy. 
Based on this modelling, the learning model could be designed to focus more on the with-in super-class differences. 
The main criteria of grouping similar action in super-class is the main verb of the action labels.
For example, super-class of below action classes is ``pretend":
\begin{itemize}
\item \texttt{Pretending to pick something up}
\item \texttt{Pretending to pour something out of something, but something is empty}
\end{itemize}

\new{As with annotations of preconditions and effect, this criteria of ``main verb" works towards pseudo labels, not ground-truth super-class.}
The number of created super-class labels is 23.

\section{Methodology}
\label{Methodology}
In this section, we elaborate the proposed framework. Since our models are constructed based on TPN, we first revisit the structure of TPN. Firstly, we propose the basic Action-Effect Combined Model where only effect was considered assuming that precondition already lies in the input images. We then develop the Cycle-Reasoning model, where each module is optimized in a cycle fashion. 
\new{The term ``module" used in this section means the part for training each type of labels (i.e., action, effect).
There are six variations of this Cycle-Reasoning model depending on type of annotations used to train the model, of which we clarify that Model(E,P,S,A) is our final version and other variations of Cycle-Reasoning model as well as Action-Effect Combined Model are parts of ablation studies, which will be discussed in~\rsec{ablation}.}

\subsection{Revisit of the Temporal Pyramid Network}
\label{Methodology.A}
\begin{figure}[tbp]
	\centering
	\includegraphics[width=0.48\textwidth]{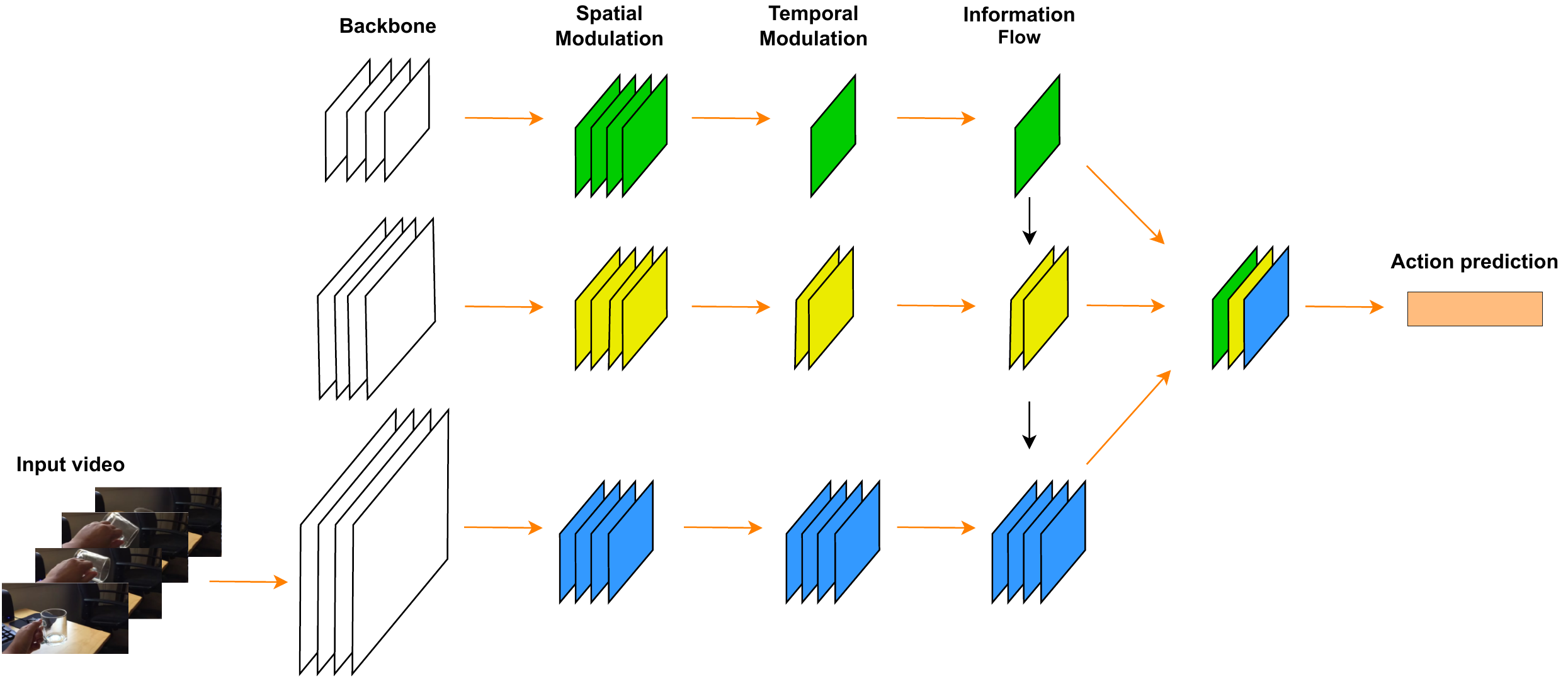}
	\caption{\textbf{Framework of TPN. } Figure reproduced from TPN~\citep{TPN}. The basic network structure of TPN was used in our models. Multiple level features extracted from the backbone are spatially and temporally downsampled by Spatial Semantic Modulation and Temporal Rate Modulation. Then features are aggregated by Information Flow in various directions and concatenated along channel dimension for final prediction. }\label{fig:figure_TPN_structure}
\end{figure}

We adopted TPN (Temporal Pyramid Network)~\citep{TPN}, which was discussed in~\rsec{Related}, as basic network structure. 
Here, we introduce the framework of TPN briefly as shown in~\rfig{fig:figure_TPN_structure}. 
The backbone, which is Resnet-50~\citep{resnet-50} with TSM~\citep{TSM}, extracts multiple level features for something-something dataset \citep{sth-sth}.
Both Spatial Semantic Modulation and Temporal Rate Modulation downsample features, but downsampling is conducted temporally in Temporal Rate Modulation. 
Information Flow collects all features in different directions.
Final Prediction part is the classification layer, concatenating pyramids of features built in previous stages.
Here, the parts corresponding to Spatial Semantic Modulation, Temporal Rate Modulation and Information Flow altogether make the the core algorithms of TPN. 
They conducted experiments with the single modality on MMAction \citep{mmaction} and initialized their models with pre-trained models on ImageNet \citep{training-imagenet}. In their model, each frame was randomly cropped as in \citep{DCNN, nonlocal}. 
They adopted augmentation of horizontal flip and a dropout of 0.5 \citep{prevent-overfitting} and did not freeze BatchNorm \citep{batch-norm}.
\new{We adopted TPN as} our baseline and basic network structure for our work due to their effective (construction of pyramids at feature-level) network with performance on the benchmark and followed this setting for our models discussed in this section. 

\begin{figure}[tbp]
	\centering
	\includegraphics[width=0.45\textwidth]{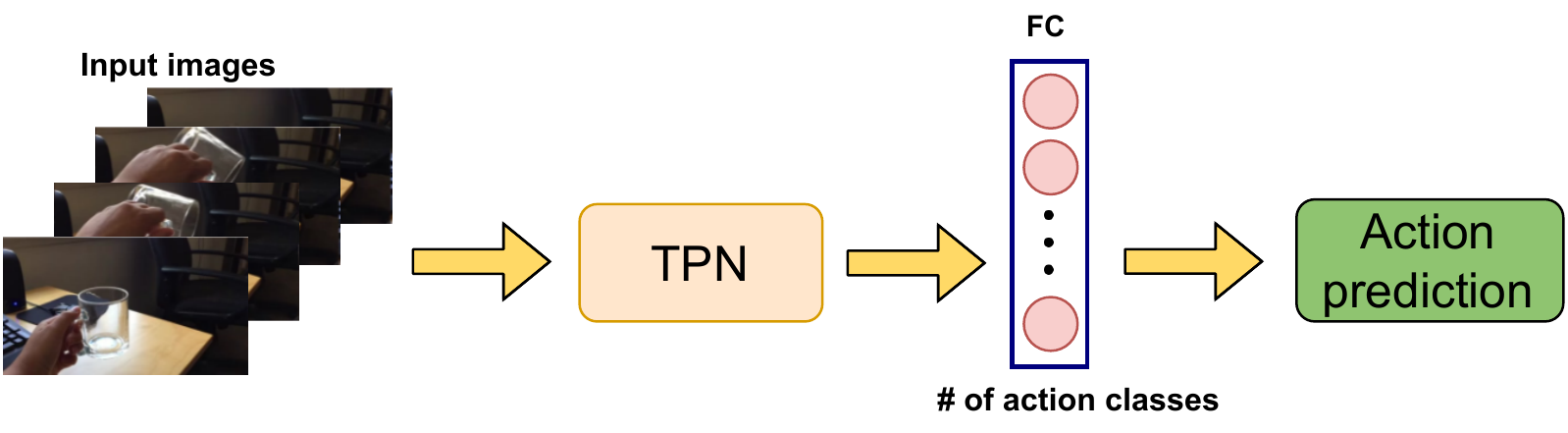}
	\caption{Action-Only model: predicts action only}\label{fig:ao}
\end{figure}

\new{\subsubsection{Action-Only Model} }
Action-Only model is the same as the TPN \citep{TPN} introduced above.
We re-implemented the model under our settings but could not reproduce the same result as TPN, so included this version as one of our baselines (See \rsec{baseline} Baseline).
For easy comparison with and understanding of other models that will be introduced in this section, structure of Action-Only model, which is just the simplified figure from~\rfig{fig:figure_TPN_structure}, is depicted in~\rfig{fig:ao}.

\subsection{Action-Effect Combined Model}
In this model, we only considered action and effect, and trained these modules simultaneously in the same network.
We developed three different structures to model the interactions between the action and the effect in different fashions.

\subsubsection{Action-Effect-Joint Model}
\begin{figure}[t]
	\centering
	\includegraphics[width=0.45\textwidth]{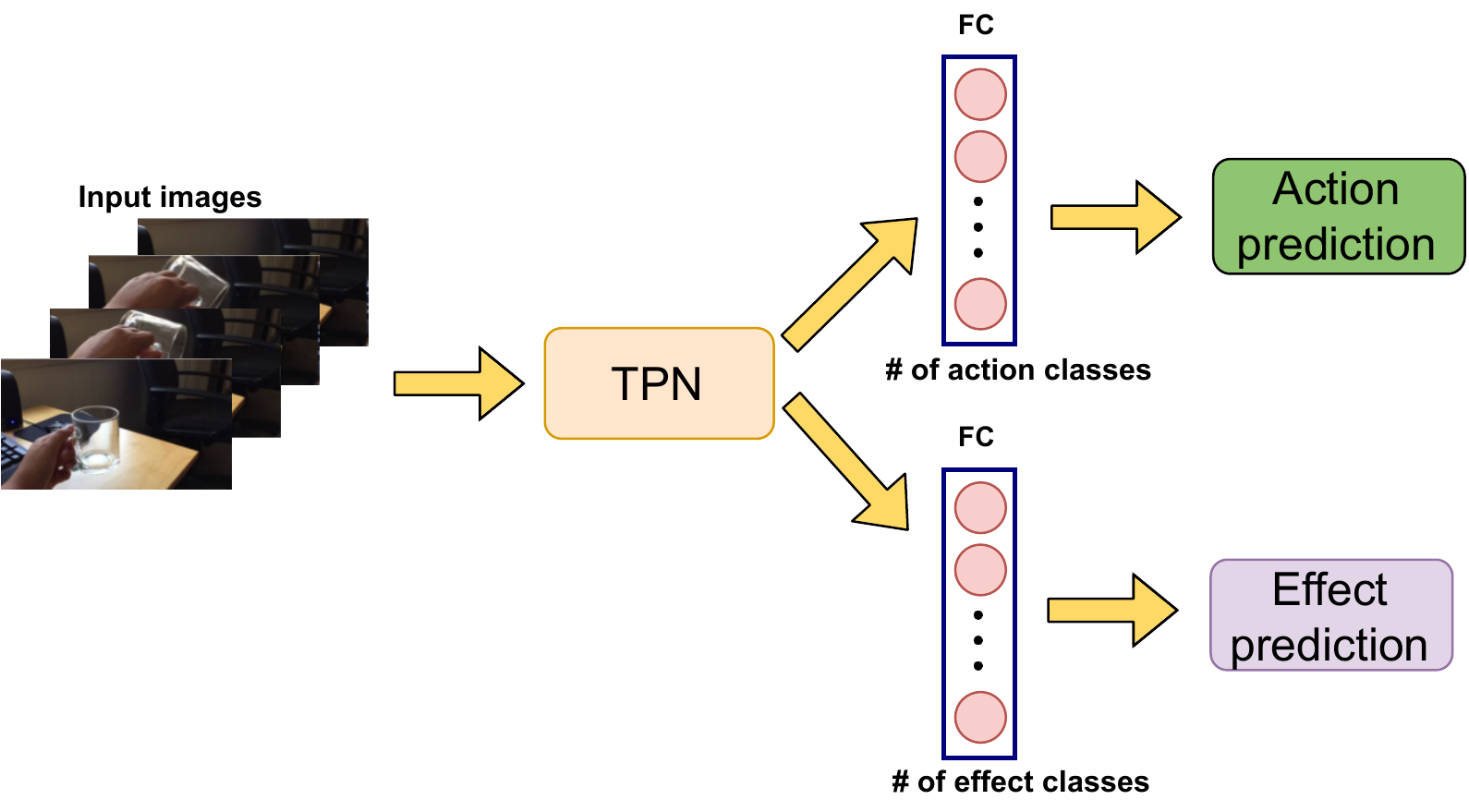}
	\caption{Action-Effect-Joint model: jointly predicts action and effect}\label{fig:jo}
\end{figure}

Action-Effect-Joint model jointly outputs action and effect. Therefore, it has two classification layers in the final layer. 
The structure of Action-Effect-Joint model is shown in~\rfig{fig:jo}.

\subsubsection{Effect-to-Action Model}
\begin{figure}[tbp]
	\centering
	\includegraphics[width=0.48\textwidth]{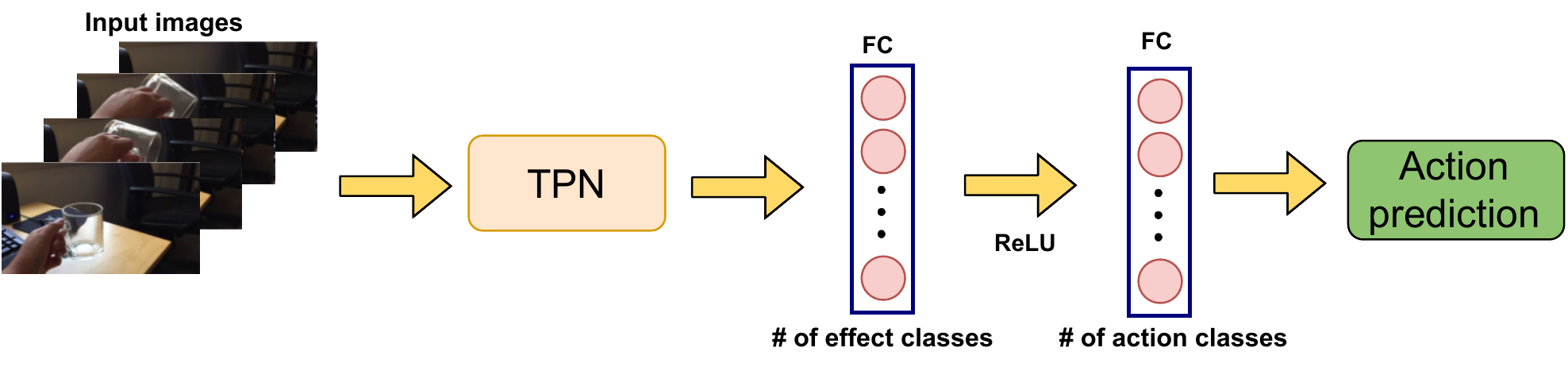}
	\caption{Effect-to-Action model: predicts action through effect}\label{fig:EA}
\end{figure}

Effect-to-Action model outputs the effect and the action successively. The model firstly outputs \new{linear logits for effect, after which ReLU activation function is applied, and softmax is used after the second FC layer} to perform action classification. This approach tries to see action through effect.
The network structure is shown in \rfig{fig:EA}.

\subsection{Cycle-Reasoning Model}
\label{Methodology.C}
\new{In our experiment, the weights for Effect module in both Action-Effect Combined Models did not work in a way to increase accuracy of action prediction.
More specifically, the weights in FC layer for Effect module after TPN branch for both 4.2.1. Action-Effect-Joint and 4.2.2 Effect-to-Action model do not contribute any better for the weight adjustment for action prediction.}
To address this limitation, we further developed the Cycle-Reasoning model.
The modules in Cycle-Reasoning model are trained separately in a cycle fashion. 
For example, Action module and Effect module in Action-Effect Combined Models discussed above are now separated from the same network and trained independently.
More specifically, in order to train action at time $T$, the softmax probabilities of classes of effect are used as inputs, which was trained at time $T-1$, so the weights for each different module (action, effect) are isolated at a specific time point in training, and the contribution to the performance from each module will not be compromised. \new{As part of ablation studies, two more modules (Precondition and Super-class) are added in Cycle-Reasoning Model.}

\begin{figure}[tbp]
	\centering
	\includegraphics[width=\columnwidth]{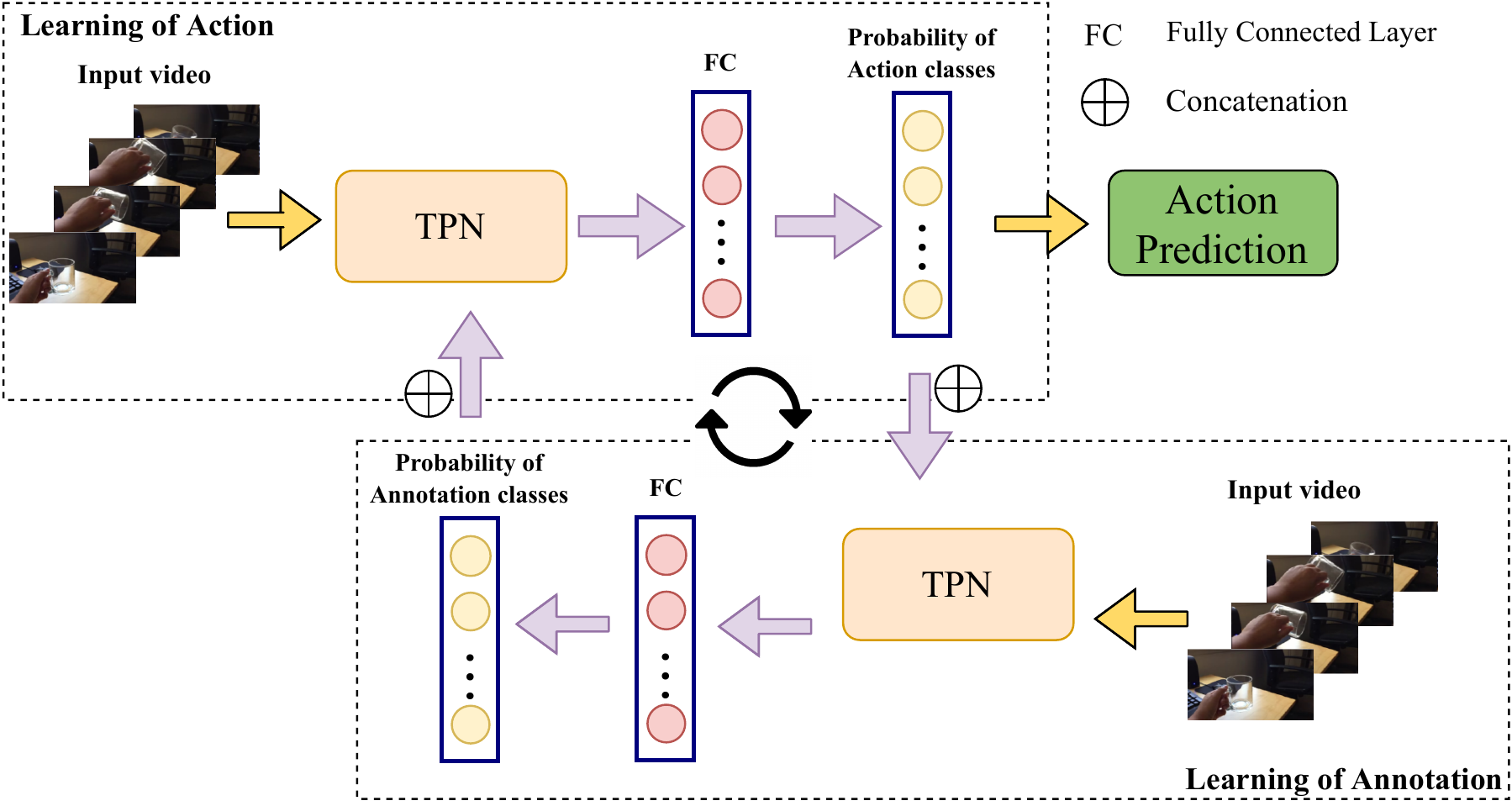}
	\caption{\textbf{Cycles for learning action using annotations.} A module for training action and a module for training annotations form a cycle. 
	The first step is to compute probabilities for all action classes as shown in the upper part of this figure (without training only for the first cycle), and train annotations with action information as shown in the bottom part of the figure. For the next step, probabilities for each annotation are computed and fed to train action model again. This cycle can be repeated as long as the improvement of accuracy is observed.}
	\label{main_cycle}
\end{figure}

An overview of Cycle-Reasoning model is depicted in~\rfig{main_cycle}, which shows the learning of different modules in a cycle. \new{Here, ``module" means the part for training each type of labels, as described earlier in this section.
The number of modules in~\rfig{main_cycle} changes depending on the kind of labels used in the bottom part of the figure. For example, if we train all of Precondition/Effect/Super-Class (Model(E,P,S) in~\rsec{ablation}), then there is 1 action prediction module in the upper part the figure, and there are 3 modules for training each annotation (Precondition/Effect/Super-Class) in the bottom part of the figure. Therefore, in this case there are 4 modules. }

The first step is to compute the prediction probability of all classes of action, and train annotations (i.e., precondition, effect) using this action prediction probability. The next step is to train action using the prediction probability of all classes of annotations trained from the previous step.
These two training processes together form one cycle. Furthermore, if the action recognition performance is better than the previous cycle, we can continue this training cycle with more iteration until the performance plateaus. As shown in~\rfig{main_cycle}, Cycle-Reasoning model consists of two parts: 1) {\itshape The Learning of Precondition/Effect/Super-class} and 2) {\itshape The Learning of Action}. We will discuss each part with more details.

\subsubsection{The Learning of Precondition/Effect/Super-class}
\label{Methodology.C.1}
As introduced above, the first step of Cycle-Reasoning model trains annotated labels, which are precondition, effect and super-class, generated in \rsec{Dataset.B}.
The key improvement from the Action-Effect Combined model  is that now the different modules are trained asynchronously. 

The network structure for training annotations is shown in the bottom part of~\rfig{main_cycle}, and each annotation (precondition, effect and super-class) is trained one at a time.
As the first step, before training annotations, we compute softmax prediction probability for all action classes with the pretrained weights of TPN \citep{TPN} for all instances of training and validation data, as shown in the upper part of~\rfig{main_cycle} (but without training action only for the first cycle when there are not models for annotations yet). 
Next, as shown in the bottom part of the figure, we train annotated labels (i.e., effect) by concatenating action class probability computed in the previous step on top of visual map features.
In the figure, TPN represents 3 core branches, which consists of Semantic Modulation, Temporal Rate Modulation and Information Flow as discussed in \rsec{Methodology.A}.
During training annotations, features extracted from images are aggregated throughout these 3 branches to form a pyramid of visual features, which then are concatenated with prediction probability of action.


\subsubsection{The Learning of Action}
\label{Methodology.C.2}
The second step of Cycle-Reasoning model trains action module in the same way as the first step above, but during training action, uses the prediction probability for all classes of annotations of side-information (precondition, effect and super-class) generated from the above step.
In other words, softmax probability for all classes of annotations trained in the last step as shown in the bottom part of~\rfig{main_cycle}, is computed for all instances of training and validation data, and this annotation prediction probability is concatenated with visual features for training action as shown in the upper part of the figure.
Step 1 (\ref{Methodology.C.1}) and 2 (\ref{Methodology.C.2}) form one cycle to predict action.
At the end of a cycle, if the accuracy is improved compared to the previous cycle, probability for each action class is computed for {\itshape The Learning of Precondition/Effect/Super-class} in the next cycle. 




\subsubsection{Training and Testing through Cycles}
\begin{figure}[tbp]
	\centering
	\includegraphics[width=\columnwidth]{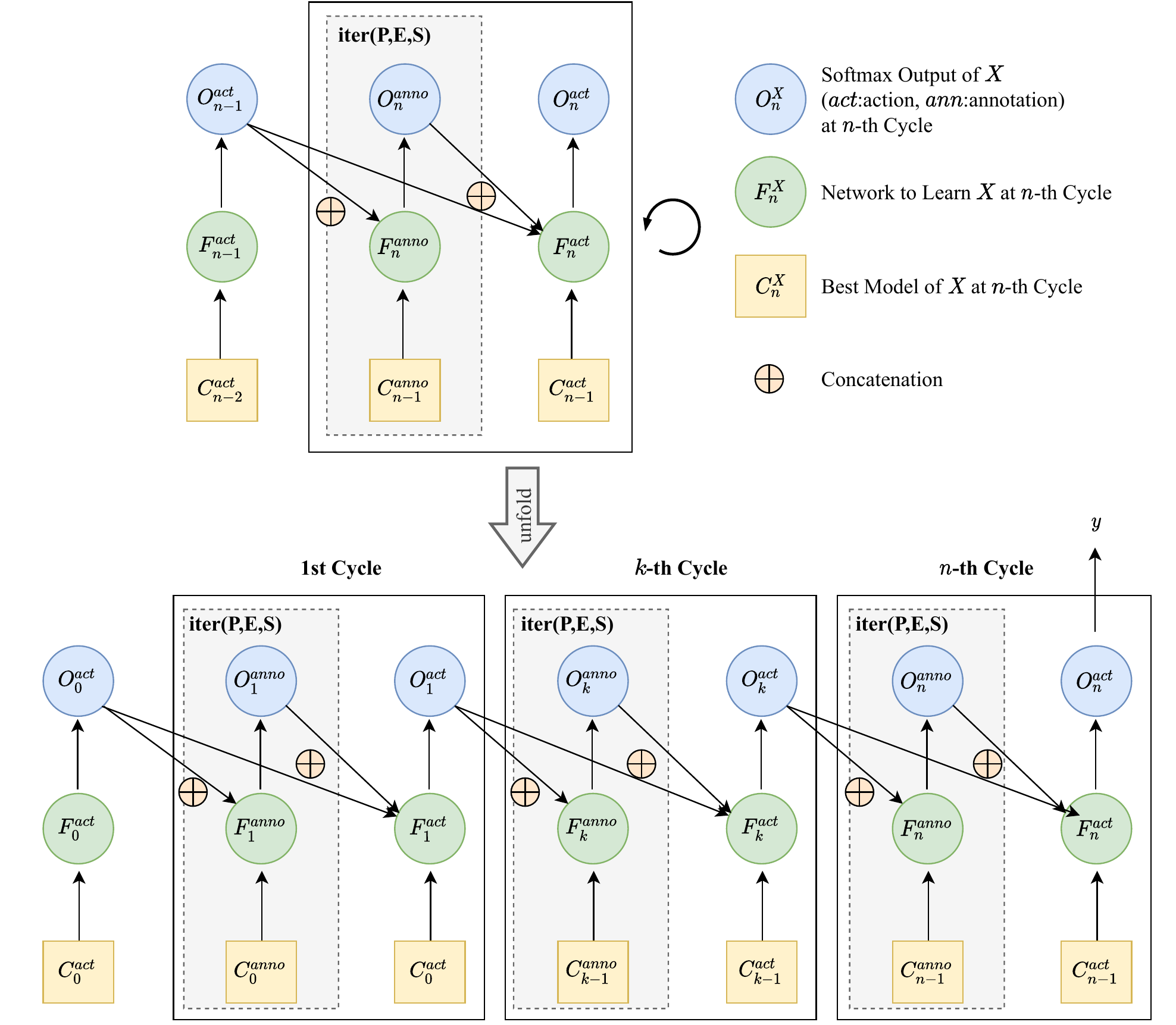}
	\caption{\textbf{Model(E,P,S,A) - Learning action through Cycles.} For training action, softmax probability of annotations from the very previous step and action softmax probability from the previous cycle are added. For training each model, best model weight saved from the previous cycle is loaded. 
    }\label{fig:cycle_unfolded}
\end{figure}

As discussed above, {\itshape The Learning of Precondition/Effect/Super-class} and {\itshape The Learning of Action} form one cycle. Training continues through cycles as long as an increase of validation accuracy is observed.
\rfig{fig:cycle_unfolded} shows the process of the training of Model(E,P,S,A) in cycles. The model is trained via fine-tuning by loading from the best weight saved in the previous cycle.
For example, in the figure, $C_{n-1}^{act}$, which is the best action model saved at cycle $n-1$, is loaded when training action model at cycle $n$. 

Testing process also follows training process except that the best model saved during the training time in Cycle $n$ is also used at the time of testing in Cycle $n$. 
For example, if the action model in Cycle 2 turns out to be the best model, at the time of testing annotation and action in Cycle 1, $C_{n-1}^{anno}$ and  $C_{1}^{act}$, which are the best annotation and action model saved during the training time in Cycle 1, are loaded respectively. In this case, testing is conducted up to Cycle 2 to predict action since Cycle 2 shows the best result at the training time.

\section{Experiments}
\label{Experiments}

\subsection{Implementation}

We trained all the models discussed in the previous section on the training data which consist of 168,913 instances, and tested on the validation data, which consist of 24,777 instances. 
When training models for the first cycle, we initially used pretrained weights of TPN \citep{TPN}, but from the next cycle, we trained the models from the best model at the previous cycle.

\new{Since visual tempo features from TPN are already powerful, we simply used FC layer for each module rather than studying more complex networks to see if adding a certain type of annotation contributes to increasing performance for action recognition.
The number of neurons in FC layer for each module is basically the sum of dimension of visual-tempo-features channel and the number of classes in each annotation added to visual-tempo features. For example, the dimension of features after TPN branch is 5 (index 0th: batch-size, 1st : visual-tempo, 2nd: time,  3rd: width,  4th: height). We concatenated the softmax values of classes for each annotation on top of visual-tempo features (index 1) from TPN. }

Following the experiment of TPN \citep{TPN}, we also used a momentum of 0.9, a weight decay of 0.0001 and a synchronized SGD training \citep{training-imagenet}, but used 4 GPUs instead of 32 GPUs that were used for TPN.
Since we used 4 GPUs, we set learning rates to 0.005 accordingly following the recommendation of TPN. Training was stopped if no increase of accuracy was shown within approximately 5--10 epochs at the time of validation. And the model saved at epoch with the best result for each model was chosen and used as final output or for training another. Spartan \citep{spartan} was used for training our model.

\subsection{Baseline} 
\label{baseline}
As discussed in the previous section, we used the basic network structure of TPN~\citep{TPN} as the baseline for performance comparison. 
We use Action-Only model of Action-Effect Combined model for this purpose.

The accuracy of the baseline is shown in~\rtab{table:table_baseline}.
First, under the same setting as TPN~\citep{TPN}, the Action-Only model, which is exactly the same network structure as TPN, \new{can achieve an accuracy of 60.75\%.}
This is our baseline1 in~\rtab{table:table_baseline}. 
Since we could not reach the reported accuracy (62.0\%) of TPN \citep{TPN}, for fair comparison, we also include the reported accuracy of 62.0\% from their paper, which is our baseline2 in~\rtab{table:table_baseline}. 

\begin{table}[tbp] 
\caption{\textbf{Baselines for action. } For action, we have two baselines. Accuracy of 62.0\% is the result of TPN reported in their paper. 60.75\% is the result that we obtained using TPN model under our environment.}\label{table:table_baseline}
  \begin{center}
  \begin{tabular}{lcc}
    \hline
    Label &baseline1 (\%) & baseline2 (\%)\\
    \hline
    Action & 60.75 & 62.00 \\
    \hline
  \end{tabular}
    \end{center}
\end{table}

\begin{table}[tbp]
 \begin{center}
  \renewcommand{\arraystretch}{1.1}
  \caption{\textbf{Prediction of Action accuracy.} Our later models outperform both baselines for action (TPN and Action-Only model).}\label{table_action_acc}
  \resizebox{\columnwidth}{!}{%
  \begin{tabular}{lc}
    \hline
    Model & Accuracy (\%)\\
    \hline
    TRN~\citep{TRN} & 48.80 \\ 
    TSN~\citep{TSN} & 30.00 \\ 
    TSM~\citep{TSM} & 59.10 \\ 
    TPN (baseline2) & 62.00 \\ 
    Action-Only (baseline1) & 60.75 \\
    Action-Effect-Joint & 61.14 \\
    Effect-to-Action & 60.14\\ 
    Model(E): Action w/ Pr(Effect) & 63.35 \\
    Model(P): Action w/ Pr(Precondition) & 63.33 \\
    Model(S): Action w/ Pr(Super) & 61.78 \\
    Model(E,P): Action w/ Pr(Eff+Pre) & 64.08 \\ 
    Model(E,P,S): Action w/ Pr(Eff+Pre+Sup) & 64.50 \\
    Model(E,P,S,A): Action w/ Pr(Eff+Pre+Sup+Act) & \textbf{65.40} \\ 
    \hline
  \end{tabular}%
}
  \end{center}
\end{table}
\new{\subsection{Ablation studies}}
\label{ablation}
In this section, we will compare the results of different models for prediction accuracy of action.

There are six settings to train the action model as follows, where ``Model(X)" means ``Action model with the softmax probability of X classes added to visual features"
\begin{itemize}
\item {\verb|Model(E)|}: Action with Probability of Effect 
\item {\verb|Model(P)|}: Action with Probability of Precondition
\item {\verb|Model(S)|}: Action with Probability of Super-class
\item {\verb|Model(E,P)|}: Action with Probability of {Effect and Precondition}
\item {\verb|Model(E,P,S)|}: Action with Probability of {Effect, Precondition and Super-class} 
\item {\verb|Model(E,P,S,A)|}: Action with Probability of {Effect, Precondition, Super-class and Action} 
\end{itemize}

The structures of these models are the same as~\rfig{main_cycle} with different types of annotations. \new{For example, for Model(E,P,S), there are actually 3 modules, that is Effect, Precondition and Super-class respectively, trained separately as shown in the bottom part of~\rfig{main_cycle} and 1 module to train action as shown in the upper part of~\rfig{main_cycle}.}
Note that in Model(E,P,S,A), probability of action in the previous cycle was included as shown in~\rfig{fig:cycle_unfolded}.

Prediction result of action for different kinds of models is shown in \rtab{table_action_acc}. The results of Model(E,P,S,A), Model(E,P,S), Model(E,P) in \rtab{table_action_acc} were trained up to second iteration of the cycle since the accuracy for action models started to drop from the third iteration.
Note that Model Action-Only, Action-Effect-Joint and Effect-to-Action are initial models, and Model(E,P,S,A), Model(E,P,S), Model(E,P), Model(E), Model(P) and Model(S) are the later models that used class probability of annotations.
Results of Two-stream  2D  CNN  methods discussed in~\rsec{Related} against something-something v2 dataset were also included for comparison in the table.
As discussed in baseline section, we have two baselines for action: 1) 60.75\% of the same model as TPN but which was trained under the same setting as all the other models in our environment (Action-Only model) and 2) 62.0\% in the paper of TPN \citep{TPN}. The best of our models, Model(E,P,S,A), outperforms TPN baseline by 3.4\% and the baseline for action obtained under our setting by 4.65\%. 

Our initial models, Action-Effect-Joint and Effect-to-Action did not show a good result compared to the baseline, but our later models except Model(S) all outperform both of action baselines. 
However, the fact that the result of action model that used annotation of super-class is lower than baseline2 (TPN) does not mean that super-class annotation is not meaningful.
Otherwise, Model(E,P,S) and Model(E,P,S,A) could not show better performance when super-class probability is added to the model. Since the accuracy that we obtained from training the same model as TPN in our environment was 60.75\% (Action-Only model), Model(S) still shows better performance than this baseline, thus annotation of super-class is better to be added as a feature. Results show that all of annotations are meaningful to improve action recognition.

\begin{table}[tbp]
   \caption{\textbf{Comparison between cycles for the learning of Precondition, Effect, and Super-class. }Here, action weights mean the action model with which to produce action class probability which is fed during training each annotation. All annotations saw improvement of accuracy by training for one more cycle, and effect has the highest accuracy in Cycle3.}\label{table_cycle_annot}
 \begin{center}
  \renewcommand{\arraystretch}{1.1}
  \resizebox{\columnwidth}{!}{%
  \begin{tabular}{llccc}
    \hline
    \multirow{2}{*}{Learning Object} & \multirow{2}{*}{Action Weights} & \multicolumn{3}{c}{Accuracy (\%)} \\
    & & Cycle1 & Cycle2 & Cycle3 \\
    \hline
    \multirow{3}{*}{Precondition} & Model(E,P,S,A) & \multirow{3}{*}{71.72} & \textbf{72.20} & 72.05 \\ 
    & Model(E,P,S) & & 72.16 & --- \\ 
    & Model(E,P) & & 72.01 & ---\\ 
    \hline
    \multirow{3}{5em}{Effect} & Model(E,P,S,A) & \multirow{3}{2em}{69.44} & 69.88 & \textbf{69.99}\\ 
    & Model(E,P,S) & & 69.87 & ---\\ 
    & Model(E,P) & & 69.77 & ---\\ 
    \hline
    \multirow{2}{5em}{Super-class} & Model(E,P,S,A) & \multirow{2}{2em}{77.55} & \textbf{77.88} & 77.87\\ 
    & Model(E,P,S) & & 77.80 & ---\\ 
    \hline
  \end{tabular}%
}
  \end{center}
\end{table}

\subsection{Further Analysis}
We discussed how model performance can be improved through repeating cycles. In this section, we discuss how model performance is changed in different cycles. 
\begin{figure*}[tbp]
	\centering
	\includegraphics[width=\textwidth]{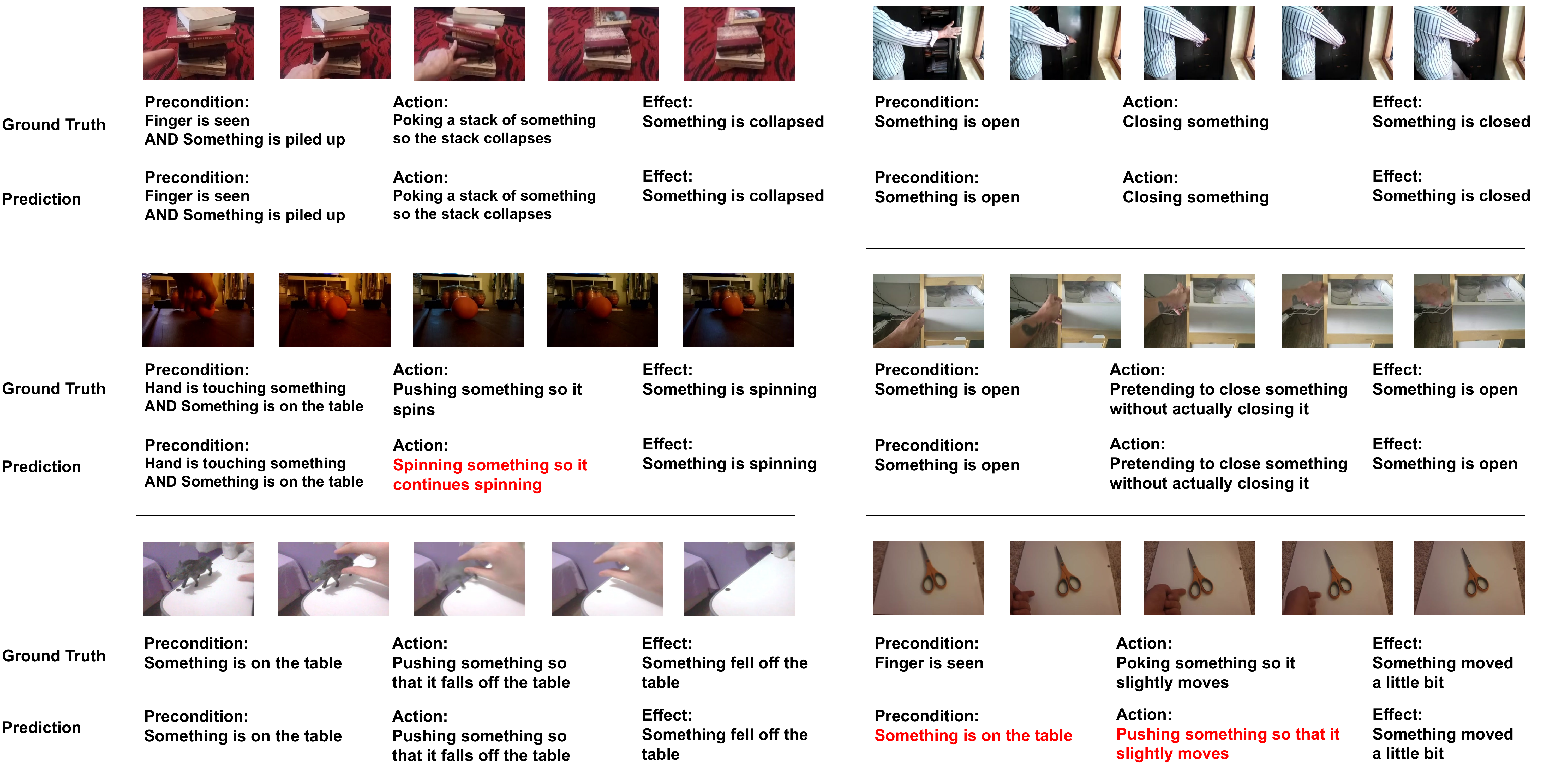}
	\caption{\textbf{Ground truth and predicted results. }}\label{figure_comparison_label_prediction}
\end{figure*}

\rtab{table_cycle_annot} and \rtab{table_cycle_action} show comparisons between cycles for {\itshape 1) The Learning of Precondition/Effect/Super-class} and {\itshape 2) The Learning of Action}.
In \rtab{table_cycle_annot} and \rtab{table_cycle_action}, the roles of the model are different. In \rtab{table_cycle_annot}, for example, when training effect, Model(E,P,S,A) in the tab of {\itshape Action Weights} means that the output of class probability of action to be fed during training effect was trained and computed with this model weight (Model(E,P,S,A)).
However in \rtab{table_cycle_action}, {\itshape Model} here means literally the kind of action model that was used for training action.

\rtab{table_cycle_annot} shows that accuracy of prediction of annotation models can be increased by training for more cycles. 
Each annotation is trained via fine-tuning by loading from the best weight saved in the previous cycle. 
For Cycle1, initially pretrained weights of TPN is used to compute action class probability to be fed during training annotations.
For Cycle2, different results are shown according to different kinds of action weights.
For Cycle3, only model weight Model(E,P,S,A) was used to compute action class probability at the end of Cycle2, which is fed during training 3 types of annotations since it showed the best performance in Cycle2. While effect has shown the highest accuracy in Cycle3, accuracy for precondition and super-class dropped in Cycle3.

\rtab{table_cycle_action} shows accuracy in different cycles for {\itshape the Learning of Action}.
As discussed above, only Model(E,P,S,A) was trained for Cycle3 since it showed the best performance. 
Although we did not see an increase in all the models compared to the previous cycle by training extra cycle, we could see an improvement of accuracy in Model(E,P,S,A) and Model(E,P,S).
However, for Model(E,P,S,A), accuracy dropped from Cycle3. Still, we saw an improvement of accuracy in Cycle2 from the previous cycle by 0.29\% in Model(E,P,S) and by 0.25\% in Model(E,P,S,A), which proves that the training for extra cycles is effective.

\begin{table}[tbp]
\caption{\textbf{Comparison between cycles for the learning of Action. } Model(E,P,S,A) and Model(E,P,S) saw an improvement of accuracy by training for one more cycle.}\label{table_cycle_action}
  \begin{center}
  \renewcommand{\arraystretch}{1.1}
  \begin{tabular}{lccc}
\hline
\multirow{2}{*}{Model} & \multicolumn{3}{c}{Accuracy (\%)} \\
 & Cycle1 & Cycle2 & Cycle3 \\ \hline
Model(E,P,S,A) & 65.15 & \textbf{65.40} & 65.29 \\
Model(E,P,S) & 64.21 & 64.50 & --- \\
Model(E,P) & 64.08 & 64.00 & --- \\ \hline
\end{tabular}
  \end{center}
\end{table}

\begin{table}[tbp]
 \caption{\textbf{Final comparison for Action. }Our final model outperforms both baselines for action.}\label{table_final_result}
  \begin{center} 
  \renewcommand{\arraystretch}{1.1}
\begin{tabular}{lc}
\hline
Model & Accuracy (\%) \\ \hline
Action-Only & 60.75 \\
TPN & 62.00 \\ \hline
Model(E,P,S,A) & 65.40 \\ \hline
\end{tabular}
  \end{center}
\end{table}

The final result for action prediction is shown in \rtab{table_final_result}. 
The best among our models, Model(E,P,S,A) outperforms both baselines of action.
\rfig{figure_comparison_label_prediction} shows examples of the comparison between ground truth and prediction result of action, precondition and effect. The red-colored text in prediction is wrongly predicted result. However, we can see that even the wrong prediction has good justification compared to the ground truth.

\section{Conclusion}
While the works on action recognition can be easily found, less importance has been placed on study of precondition and effect. However, the study of their relationship reasoning in action is of significant importance since it plays an important role in reasoning their relationship with action (precondition-action and action-effect), and action recognition can be improved and robust in generalization in the relationship with precondition and effect. In this context, study on the recognition of precondition and effect is very meaningful.

In this paper, we create labels of precondition, effect and super-class based on the action labels provided from the datasets. Next, we discussed different models to predict annotations and then action by using these annotations. We showed that 1) prediction of annotations (i.e., effect) can be improved by using class probability of action and 2) that these annotations relevant to actions in turn can be used to improve accuracy of action recognition. 

Without modifying the network to a large extent, we proved that a significant improvement can be achieved from the existing network simply by using annotations relevant to action. Furthermore, this approach can be applied to any network. \new{This finding could be further developed in later experiments by using more sophisticated methodology to extract ground truth for precondition and effect rather than relying on manual annotations. The approach used in this paper is useful for future studies on the action recognition in the context of precondition and effect.}

\bibliographystyle{model2-names}
\bibliography{refs}

\end{document}